# evt_MNIST: A spike based version of traditional MNIST

an event-based MNIST

Mazdak Fatahi
Computer Engineering Department
Razi University
Kermanshah, Iran
Mazdak.fatahi@gmail.com
Tel Number: +98 9183592337

Mahyar Shahsavari
CRIStAL laboratory
University of Lille
F-59000 Lille, France
Mahyar.Shahsavari@ed.univ-lille1.fr

Mahmood Ahmadi
Computer Engineering Department
Razi University
Kermanshah, Iran
m.ahmadi@razi.ac.ir

Arash Ahmadi
Electrical Engineering Departmentine
Razi University
Kermanshah, Iran
A.ahmadi@razi.ac.ir

Philippe Devienne
CRIStAL laboratory
University of Lille
F-59000 Lille, France
Philippe.Devienne@univ-lille1.fr

*Abstract*— **Benchmarks and datasets have important role in evaluation of machine learning algorithms and neural network implementations. Traditional dataset for images such as MNIST is applied to evaluate efficiency of different training algorithms in neural networks. This demand is different in Spiking Neural Networks (SNN) as they require spiking inputs. It is widely believed, in the biological cortex the timing of spikes is irregular. Poisson distributions provide adequate descriptions of the irregularity in generating appropriate spikes. Here, we introduce a spike-based version of MNSIT (handwritten digits dataset), using Poisson distribution and show the Poissonian property of the generated streams. We introduce a new version of evt_MNIST which can be used for neural network evaluation.**

I. INTRODUCTION

There are different databases to verify the accuracy and performance of machine learning algorithms. These databases are presented in various fields. For example to evaluate neural networks recognition rate of facial expressions. A lot of training data are required to test and validation. There are some widely used standard databases that become known to compare the results to similar works. However all known databases are applicable for Artificial Neural Networks (ANN). Consequently, this challenge is remained in usage of traditional frame-based images in existing datasets. Spiking Neural Networks use spikes for communication between the neurons. In this communication, the spikes are the same, a spike by itself will not carry any information and the number and the timing of spikes are important [1]. Therefore, if we want to use these benchmarks to evaluate our works, we need to extract proportional spike sequence from them, considering, the proper number of spikes and interval between them.

There are limited public method alternatives to make a suitable dataset to check the validation of SNN. Indeed if it is not possible to use these public methods, we need to transfer the ANN dataset to Spiking one. For instance in [2], each image is presented to the network for 350ms in the form of a spike stream with Poisson-distribution. To encode the input images to the spike trains, firing rates has been transferred between 0Hz and (255/4)Hz. They repeat this process until at least five spikes have been fired during the presentation time. Authors of [3, 4], presented the input as asynchronous voltage spikes using some coding approaches. To convert the static images into events stream (spike train), spikes are generated with probability proportional to the intensity of the pixel of a given image.

According to our investigations, in many perfect SNN architectures [5-8] especially in image processing application, the stimuli is generated directly from a spiking "retina" that naturally presents data as asynchronously. In [7] to response to the visual inputs, the spikes are produced by the Dynamic Vision Sensor (DVS). Furthermore a subset of the emitted spike by the DVS (randomly), are mapped into hidden layer neurons.

The Neuromorphic vision sensors are not publicly available. A spike train is a sequence of spikes generated by a single neuron. A spike train can be defined as a sequence of spike times. This chain of events can occur at regular or irregular intervals. Some evidence for neuronal variability and spike-train irregularity were reviewed in [1]. In the biological cortex, action potential timing is irregular and it is not periodic. Each irregular spike sequence can be consider as a stochastic sequence generated by a Poisson process. Here, we assume that the generation of each spike is independent of all other spikes. Respecting to this in dependency, the spike train would be completely described through a Poisson process. It is notable that, some aspects of neuronal dynamics can break the independent spike assumption [9]. In addition, we postulate that the firing rate for each synapse is constant over time. This means we assume the Poisson process is homogeneous. Indeed Poisson process is the simplest stochastic description of neuronal firing. As it was shown in [10], since each spike in a Poisson process is independent, then Poisson firing cannot be taken into account for refractoriness. The using of renewal processes, which is depends on the time of the last spike is suggested. Practical works have shown the Poisson process is adequate for spike generation and evaluation the proposed algorithms.

Here we used Poisson distribution to generate spikes train of images of MNIST dataset to extract a spike-based dataset from a frame-based dataset respecting to intensity of each pixel of images.

Accordingly, in the next section we will introduce Dynamic Vision Sensor (DVS) and MNIST dataset briefly. We present our proposed algorithm and talk about evt_MNIST as a ready to use spiking version of MNIST in section III. Finally we evaluate the accuracy of the proposed method for spike generation as a Poisson sequence in section IV and in section V, the paper is concluded.

## II. INTRODUCING DYNAMIC VISION SENSOR (DVS) AND MNIST

### A. Dynamic Vision Sensor (DVS)

"Retina" as a common vision sensor in Neuromorphic domain, is an Address-Event Representation (AER) bio-inspired image sensor which measures visual information based on spikes train from scenes and generates corresponding spike sequences. AER is a Neuromorphic communication protocol for spiking neurons. In AER, spikes are represented as digital addresses and leading to the address-event representation of images. In [11] an Asynchronous Temporal Contrast Vision Sensor was proposed which can independently generate spike events from each pixel relative intensity changes. Dynamic Vision Sensor (DVS)( Fig. 1), used as Asynchronous Temporal Contrast Vision Sensor and works similar to retina. Instead of sending entire images, only the changes of local pixel-level and the time of changes are transmitted (Fig. 2).The output of the sensor a is stream of events at microsecond time resolution.

These Neuromorphic vision sensors are not publicly available for all researchers. Indeed it was our motivation to exposing evt_MNIST as a public spiking benchmark.

### B. MNIST: database of handwritten digits

MNIST database (Mixed National Institute of Standards and Technology database)([12]) is a widely used database for testing the network recognition rate of many neural network related research. It becomes a standard way of comparing different works in neural network domain. For example if we need to evaluate a new algorithm of learning for recognition of facial expressions, we need a lot of train and test data that MNIST provide us.

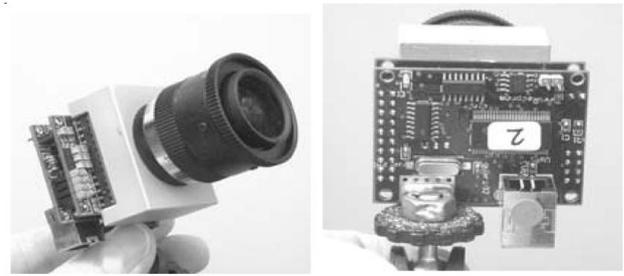

Fig. 1: Dynamic Vision Sensor (DVS) - asynchronous temporal contrast silicon retina

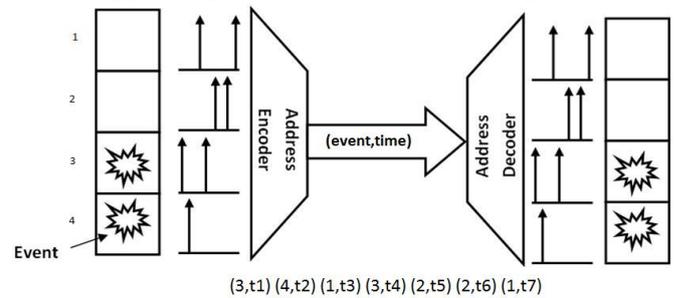

Fig. 2: Encoding and Decoding process according to AER protocol

To compare the work with others, a database similar to others is required. For this problem there are standard databases. One of this databases in handwritten digits recognition is MNIST.

MNIST dataset has been widely used as a benchmark for verify and evaluating classification algorithms in handwritten digit recognition systems. For example in [13], professor Hinton to show the accuracy of Deep Belief Networks (DBN) to compare with Virtual SVM, Nearest Neighbor and Back-Propagation used MNIST database. In [2, 4, 14-16] MNSIT is used for evaluation the proposed approaches.

The MNIST is widely used for training and testing in the field of machine learning. In some papers the training set was extended by adding some distortions (random combinations of scaling, shifts, adding some noise etc.).

The MNIST database is available on MNIST webpage [12] and contains of 60,000 examples for training set and 10,000 examples for test set. The technical issues are described in our report [17]. Fig. 3 illustrate some exported images from MNIST.

## III. GENERATING AN EVENT DRIVEN VERSION OF MNIST

According to our introduction, here we used Poisson distribution to generate spikes trains of images of MNIST dataset to extracting a spike-based dataset from a frame-based dataset respecting to intensity of each pixel of images. There are two commonly used procedures for generating homogeneous Poisson spike trains [10]. The first one is driven from the relevance of Poisson and Exponential distributions. If we choose Inter Spike Intervals (ISI) randomly from a given exponential distribution, cumulating these ISIs can generate continues sequence of spike times.

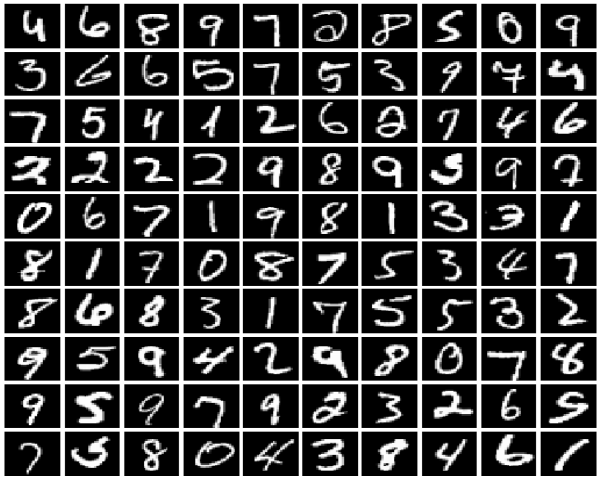

Fig. 3. 100 digits from MNIST

According to our assumption (the generation of each spike is independent of all the other spikes), we also assume the spike generation is only depended on instantaneous firing rate.

In our simulation, we postulate any pixel's density of MNIST digits, as the instantaneous firing rate of stimuli is clamped on inputs neurons during the simulation time. Indeed we refer to intensity of pixels as the probability that a spike occurs within an interval. Then we can use the MNIST digits vector as a probability vector. If we divide the simulation time into time periods $\Delta T$, which in each period maximum number of spike is 1, then we have n bins:

$$Bins_{Number} = \frac{Simulation_{Time}}{\Delta T} \quad (1)$$

The firing rate r (pixel's density) determines the probability of firing a single spike in any $\Delta T$. The probability of occurring a single spike in any $\Delta T$ is:

$$r\Delta T \quad (2)$$

In homogeneous Poisson process, rate of spike generation in any $\Delta T$ is only depended on stimuli intensity in $\Delta T$ respecting to constant intensity of pixels during the simulation, for all intervals of simulation time ($\Delta T$), we have constant spike generation probability as the mean firing rate.

Therefore according to this fact that the probability of firing a spike during a short period of time (i.e.: 1ms) is $r\Delta T$, using a simple sampling procedure (Algorithm 1), each generated spikes will be assigned to discrete time.

ALGORITHM 1: POISSON SPIKE TRAIN GENERATION

```
1: Find M as the number of bins:
       M = Simulation_Time / ΔT
2: Generate M uniform random numbers between 0:1
   as sequence:
       X = [x_rand]
3: For each bins:
       if (rΔT > X_i)
           generate 1 spike
       else
           do nothings
```

It has been proven in [18], when the probability of having 1 spike in $\Delta T$ is $r\Delta T$, then the probability of having n spikes in total time ($T$), $P_T[n]$, is a Poisson distribution. Using this method, we generate an event-driven dataset of MNIST called evt-MNIST[1] [19].

In [19] two videos can be found which playback the spike generation corresponding to pixels rate. You can see pixels with more density (the righter pixels) have more spikes/sec. We assumed 100 ms for pixels representation. Indeed we used the pixels density as the intensity of stimuli which is clamped on the inputs of our SNN during of simulation. The maximum rate for completely white pixels is 1000 Hz (100 spikes in 100 ms).

In fact we prepared an AER-like data set which is consisted of 60000 matrix for training images and 10000 for test in size of 784*100 for each matrix. We assume each pixels of each pixels as an input stream for triggering the corresponding neuron in our proposed architecture in next chapter.

Each input stream is a spike train which is generated using described algorithm. Therefore we have a matrix of 784*100 logical variables. For each pixel, we have a 1*100 spike train of true and false (true = spiked /false=no things), which as it is shown before, the number of spike in each trails have Poisson distribution. It is clear that for a pixel which is brighter, more "true" in spike train is generated. In Fig. 6, we can see the generated spikes for a sample image. When we use these generated spikes train as input for a Spiking Neural Net, the neurons which receive this streams of more spikes, will be more triggered. Fig. 4 depicts the effect of number of spikes which is directly depended on pixels density and the presentation time of each image. We can see the pixels with lower intensity in contrast to pixels with higher value in the same time are less bright.

Similarly in Fig. 5, we can see the effect of randomness in spike generation. Indeed pixels with higher value have more spikes and can be seen mostly "on". The pixels with middle value, proportional to their pixel density are "on" and "off" time to time, but pixel with very low density are mostly "off". This issue is illustrated in the raster-plot of spike trains of 784 neuron (corresponding to number of pixels in a MNIST image) in 1000 ms (Fig. 7). The mostly "on" pixels are located in the center of images, consequently we can see denser spike trains in the middle of Fig. 7.

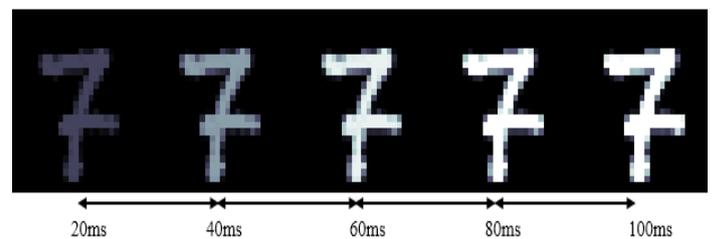

Fig. 4. Effect of number of spikes: When an input triggered continuously, the receptor neurons can be more determinative in learning process

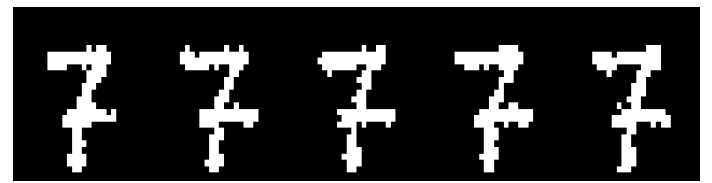

Fig. 5. The effect of pixel's density in spike generation

---

[1] We have converted all the test and train images of MNIST dataset and shared it on Internet:
https://github.com/MazdakFatahi/evt-MNIST.

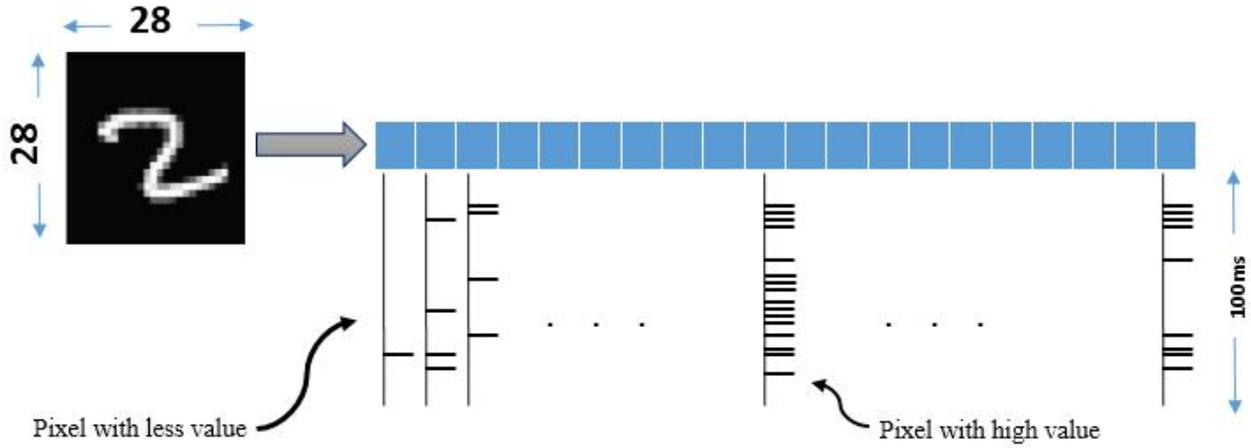

Fig. 6. Converting static images to dynamic spike trains: It's clear the pixels with less value have less corporation in learning process then in spike t streams the have les spikes and conversely about pixels with high value.

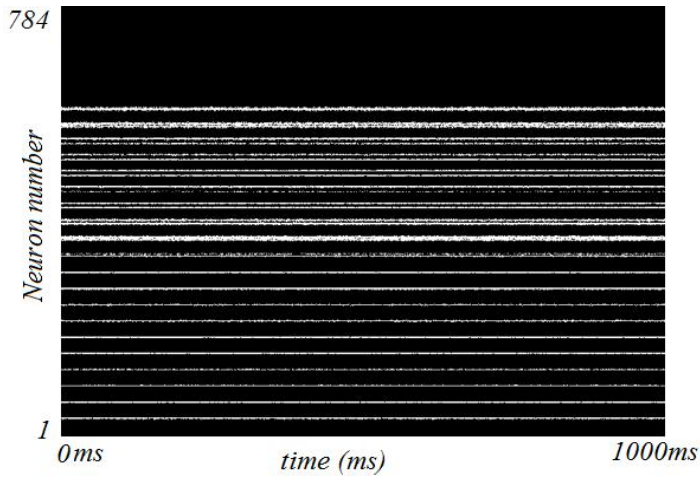

Fig. 7. Raster-plot of spike trains for a given image

## IV. METHOD EVALUATION

To evaluate the accuracy of the proposed method for spike generation in Poisson distribution, respecting to definition of Poisson process, we have to setup a simulation. In this simulation we generate some spike trains from a given pixel density, which is equal to sampling a specific neuron in a certain time interval for many trials. Using Algorithm 1, we can generate spikes and count the number of spike in each trial. Finally we have to evaluate the given sequence of firing numbers in different trials:

*Number of Spikes in k trials=$\{ n_1, n_2, n_3,\ldots , n_k\}$*     (3)

One of the best measurements for evaluating a given distribution to see if it is a Poisson process is Fano Factor (4). Since for a Poisson distribution variance and mean of the event count are equal, then the Fano Factor for a Poisson process is one.

$$FanoFactor = \frac{Var_{Dist.}}{Mean_{Dist.}} \quad (4)$$

Fig. 8 shows a chosen digit from MNIST for spike train extraction using the Algorithm 1. The instantaneous firing rate is chosen from a pixel of the vector containing the pixels value of Fig. 8. We repeat the spike generation 1000 times and generate 1000 spikes trains. Counting the number of spikes in each trail, we have a vector of 1000 numbers, which we have to estimate its distribution and Fano Factor. Using Matlab for simulation, Fig. 9 shows the histogram of the given vector and the fitted distribution. The estimated Poisson λ_hat [2] parameter (is equal to r) using Matlab is 119.569 which is very close to the chosen pixel density. We can find other parameter in Tabel 1. As it is obvious in Table 1, the estimated Fano Factor is close to 1.

We also use Easyfit software to fit probability distributions to the given vector of numbers Fig. 10.

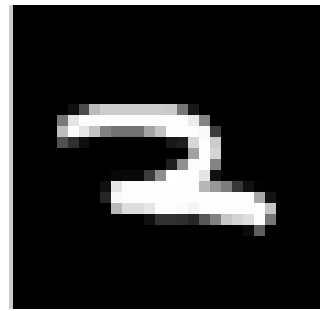

Fig. 8. One MNIST digit for extacriting corespoding spike trains for each pixels

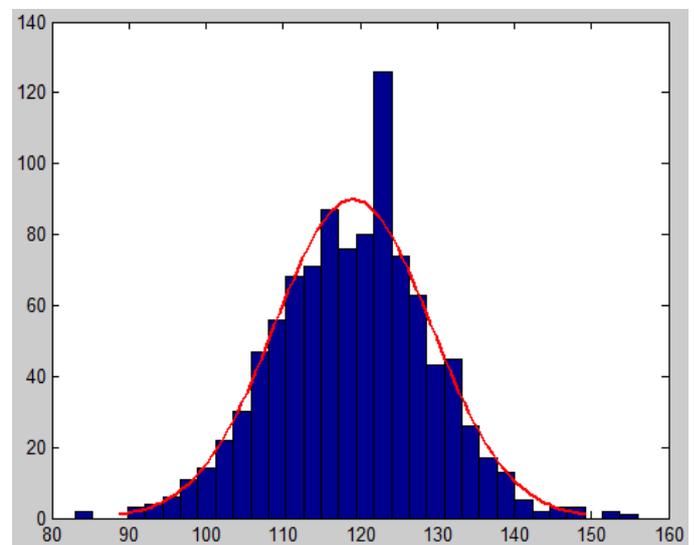

Fig. 9. Histogram of spike train simulation

---

[2] λ_hat is the maximum likelihood estimate (MLE) of the parameter of the Poisson distribution, λ, for the given data.

TABLE 1. ESTIMATED PARAMETERS FOR THE SPIKE TRAIN SIMULATION

| Estimation Parameter | Estimated value |
|---|---|
| $\lambda\_hat$ | 119.569 |
| *Variance* | 112.9142 |
| *Mean value* | 119.5690 |
| *Fano Factor* | 0.9443 |

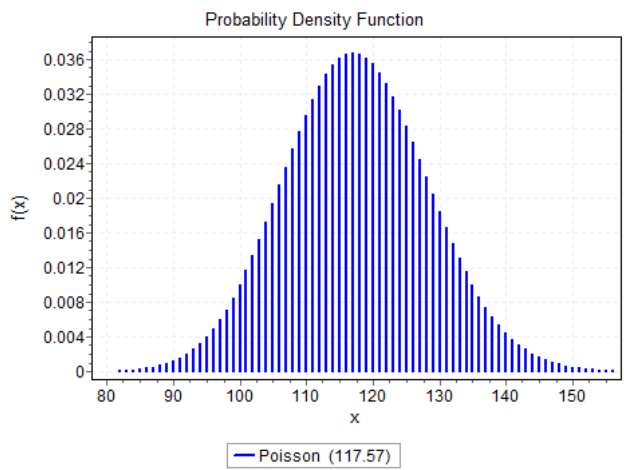

Fig. 10: Easyfit distribution estimation

## V. COCLOSIONS

Because there is no suitable dataset for evaluating SNN models, as well as lack of public availably of of the specific hardware (such as DVS), in this paper we report our proposed event-driven MNSIT called evt_MNIST. evt_MNIST is a prepared dataset which is consisted of all MNIST test and train digits sets. We prepare our dataset in Matlab file format which can be used in Brian [20] and Matlab. Corresponding to each image, we have a matrix containing 100 ms spikes train of all the pixels of each given image. Indeed maximum firing rate for each spikes train is 1KHz. We have shown the generated spike sequences have Poisson distribution which is the adequate model for simulating spikes irregularity in cortex.

## ACKNOWLEDGMENT

We would like to thanks Daniel Neil for his helpful comments. This research has been supported by Center of International Scientific Studies and Collaboration (CISSC) of Iran as Gundishapour project.